\documentclass[lettersize,journal]{IEEEtran}
\usepackage{amsmath,amsfonts}
\usepackage{algorithmic}
\usepackage{algorithm}
\usepackage{array}
\usepackage[caption=false,font=normalsize,labelfont=sf,textfont=sf]{subfig}
\usepackage{textcomp}
\usepackage{stfloats}
\usepackage{url}
\usepackage{verbatim}
\usepackage{graphicx}
\usepackage{bbding}
\usepackage{colortbl}
\usepackage{newfloat}
\usepackage{colortbl}
\usepackage{booktabs}
\usepackage{pifont}
\usepackage{xcolor}
\usepackage{newfloat}
\usepackage{amsfonts}
\usepackage{multirow}
\usepackage{multicol}
\usepackage{listings}
\usepackage{comment}
\usepackage{cite}
\begin{document}

\title{AUCH-Net: Action Unit-Based Consistency-Aware Hypergraph Network for Cross-Domain Few-Shot Facial Expression Recognition}

\author{Xinhan Qiu,
        Yan Yan,~\IEEEmembership{Senior Member,~IEEE,}
        Rui Zhu,
        Si Chen,~\IEEEmembership{Senior Member,~IEEE,}
        \\
        Hanzi Wang,~\IEEEmembership{Senior Member,~IEEE,}
        % <-this % stops a space
        % \thanks{Manuscript received April 19, 2021; revised August 16, 2021.}% 按需保留

\thanks{
This work was supported in part by the National Natural Science Foundation of China under Grant 62372388, Grant U21A20514 and Grant 62571466, the Major Science and Technology Plan Project on the Future Industry Fields of Xiamen City under Grant 3502Z20241029 and Grant 3502Z20241027, the Fundamental Research Funds for the Central Universities under Grant 20720240076 and Grant ZYGX2021J004, and the Foundation of Fujian Key Laboratory of Pattern Recognition and Image Understanding under Grant PRIU25-02.
 \textit{(Corresponding author: Yan Yan.)}}

\thanks{Xinhan Qiu, Yan Yan, and Hanzi Wang are with the
 Key Laboratory of Multimedia Trusted Perception
and Efficient Computing, Ministry of Education of China, School of Informatics, Xiamen University, Xiamen 361102, China, and also with the Fujian Key Laboratory of Pattern Recognition and Image Understanding, Xiamen University of Technology, Xiamen 361024, China  (e-mail: qiuxinhan@stu.xmu.edu.cn; yanyan@xmu.edu.cn; hanzi.wang@xmu.edu.cn).}

\thanks{Rui Zhu is with the Bayes Business School, City St George’s, University of London, London, EC1Y 8TZ, UK (e-mail: rui.zhu@city.ac.uk).}

\thanks{Si Chen is with the School of Computer and Information Engineering, Xiamen University of Technology, Xiamen 361024, China (e-mail: chensi@xmut.edu.cn).}
}

\markboth{Journal of \LaTeX\ Class Files,~Vol.~14, No.~8, August~2021}%
{Shell \MakeLowercase{\textit{et al.}}: A Sample Article Using IEEEtran.cls for IEEE Journals}


\maketitle

\begin{abstract}
Recently, cross-domain few-shot facial expression recognition (CF-FER) has received considerable attention. However, the performance of existing CF-FER methods is still unsatisfactory due to inferior transferable feature learning 
under large domain discrepancy and limited target samples. 
Fortunately, the action units (AUs), which indicate the movements of different facial muscles, provide consistent conceptual semantics for describing expressions within and across domains.
Inspired by this, we propose a novel 
\textbf{A}ction \textbf{U}nit-based \textbf{C}onsistency-aware \textbf{H}ypergraph \textbf{Net}work (\textbf{AUCH-Net}), which constructs
consistency-aware hypergraphs on AUs, for CF-FER. 
Specifically, AUCH-Net presents a new AU feature learning (AFL) module and a new visual feature learning (VFL) module. 
The AFL module learns AU features under the guidance of a novel relation consistency loss and an AU regularization loss, while the VFL module learns visual features supervised by a relation consistency loss and a classification loss.  
By learning consistent AU features, AUCH-Net 
effectively models the connections between AUs and expression categories. 
As a result, we can bridge the gap between fine-grained facial variations and high-level expression categories, greatly facilitating the learning of transferable feature representations.
Extensive experiments on both in-the-lab and in-the-wild datasets show that our method consistently outperforms several state-of-the-art methods. Our results clearly show that modeling the relationships among AUs holds significant potential for FER under cross-domain few-shot scenarios.
\end{abstract}

\begin{IEEEkeywords}
Facial expression recognition, cross-domain few-shot learning, action unit, hypergraph modeling.
\end{IEEEkeywords}

\section{Introduction}
\label{sec:intro}

% introduction
\IEEEPARstart{O}{ver} the past few decades, automatic facial expression recognition (FER) has attracted considerable attention due to its widespread applications in human-computer interaction, psychological monitoring, and health diagnostics~\cite{ jiang2012brain}. 
% basic fer
A variety of FER methods ~\cite{ ruan2021feature, wang2020suppressing, zeng2022face2exp} have been developed and achieved promising performance. These methods mainly focus on the classification of basic expressions (including happy, surprised, angry, sad, fearful, and disgusted) according to Ekman and Friesen's work \cite{ekman1971constants}. 
% compound fer
Recent research \cite{Du_Tao_Martinez_2014} reveals that 
humans regularly use many more facial expressions to convey emotions than basic expressions. To comprehensively understand facial expressions, Du \emph{et al.}~\cite{Du_Tao_Martinez_2014} define compound expressions, which are constructed as meaningful combinations of basic expressions. For example, the happily-surprised expression can be viewed as a combination of happy and surprised expressions. 
Compared with basic expressions, compound expressions include a larger number of categories and can better describe human emotions. 

\begin{figure}[!t]
    \centering    \includegraphics[width=0.46\textwidth]{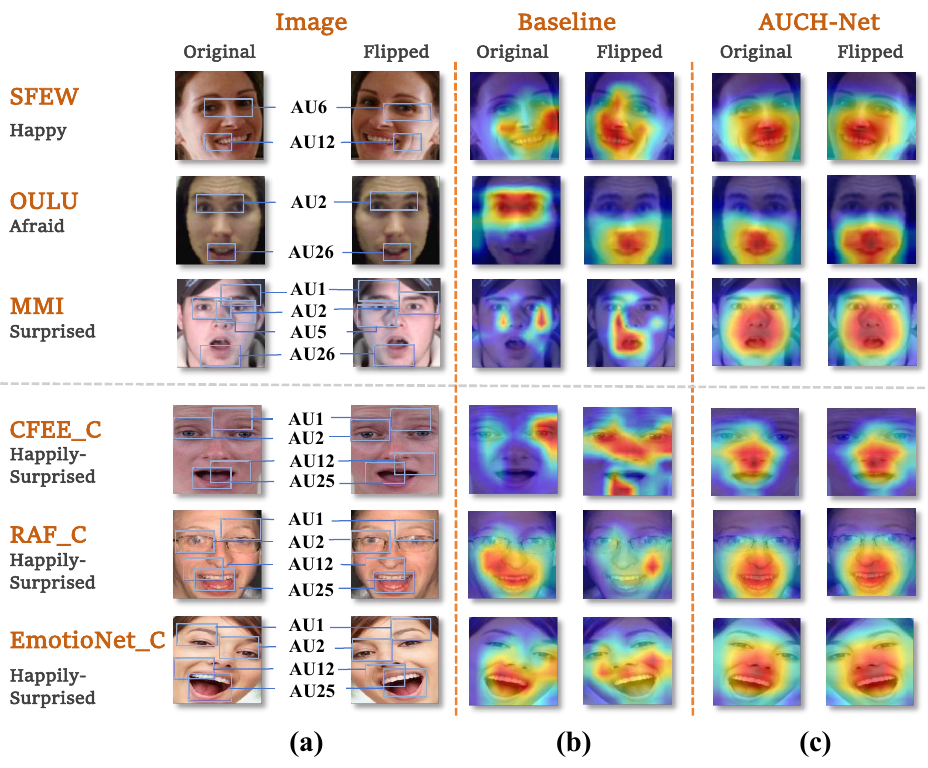}
    \caption{{Visualization of AUs and heatmaps in the source domain (three upper rows: basic expression datasets) and the target domain (three lower rows: compound expression datasets).
    % 原文
     (a) shows the expression images, their AUs, and their flipped counterparts. (b) shows the inconsistent heatmaps obtained by  Baseline \cite{tian2020rethinking}. (c) shows the consistent heatmaps obtained by our AUCH-Net.}
    }
    \label{fig:intro}
\end{figure}

% problem
Unfortunately, the manual annotation of compound expression images usually requires the guidance of psychologists due to subtle differences between compound expressions. Subsequently, annotating a large amount of training data is time-consuming and labor-intensive. 
% fsl
To reduce the expensive annotation cost, cross-domain few-shot FER (CF-FER), which identifies novel compound expressions (involving only a few samples) in the target domain by using the model trained only on easily accessible basic expressions in the source domain, has emerged as a promising learning scheme. 
% 许多基于CF-FER的methods取得了巨大的进展，Zou 等人develop a two branch的EGS-Net实现episodic learning和batch learning的交替学习，同时学习表情特征和对比能力。进一步的，Zou 等人提出CDNet其利用cascading parameter-shared decomposition module实现特征级别分解。Chen等人使用yperbolic space-based mixture-of-experts (MoE) layersand a Hyperbolic Self-Paced Learning (HSL) strategy to model hierarchical expression relationships.
Some CF-FER methods have been developed and have achieved remarkable progress. Zou \emph{et al.}\cite{zou2022facial} develop a dual-branch emotion guided similarity network (EGS-Net) to alternatively perform episodic training and batch training, learning facial expression features with good generalization ability. Later, Zou \emph{et al.}\cite{zou2022learn} propose a cascaded decomposition network (CDNet) which utilizes multiple cascaded parameter-shared decomposition modules to achieve feature-level decomposition. Chen \emph{et al.}\cite{chen2025hyperbolic} introduce HSM-Net, which employs hyperbolic space-based mixture-of-experts (MoE) layers and a hyperbolic self-paced learning (HSL) strategy to model hierarchical expression relationships.

% their problems (totally updata)
Despite great progress in recent years, the performance of existing CF-FER methods is still far from being satisfactory. On the one hand, these methods 
directly learn visual representations from fine-grained facial variations in basic expressions. Consequently, they fail to extract discriminative features for identifying novel compound expressions because of the susceptibility of visual representations under large domain discrepancy and limited target samples. 
As shown in Fig.~\ref{fig:intro}(a), the facial images from the source and target domains show large differences.

On the other hand, 
these methods only extract pixel-wise or region-wise visual representations from images. They do not explicitly consider high-order structure information, which indicates richer and more consistent relationships (e.g., the happy expression involves the actions from the eyes, eyebrows, and mouth; the left eye and the right eye exhibit similar actions for one expression). 
Hence, they struggle to extract consistent feature representations across source and target domains.  As shown in Fig.~\ref{fig:intro}(b), the heatmaps (for a facial image and its flipped counterpart) obtained by the Baseline  \cite{tian2020rethinking} significantly differ, a pattern occurring in both source and target domains.

% our method (totally update)

 Based on the above analysis, existing CF-FER methods are difficult to extract \textit{ highly transferable feature representations} under the cross-domain few-shot settings. To address the above issue, we propose a novel \textbf{A}ction \textbf{U}nit-based \textbf{C}onsistency-aware \textbf{H}ypergraph \textbf{Net}work (\textbf{AUCH-Net}), which constructs consistency-aware hypergraphs on action units (AUs) to learn consistent AU features, for CF-FER. 
Our method is inspired by facial action coding systems (FACS) \cite{ekman1978facial}, which define a set of AUs. Each AU codes the fundamental movements of individuals/groups of muscles in the face, providing consistent conceptual semantics across basic and compound expressions (e.g., the AUs of one expression category are the same 
as shown in Fig.~\ref{fig:intro}(a)). Hence, once a model learns the connections between AUs and expression categories, it can be readily adapted to infer novel compound expressions with limited samples (by leveraging conceptual AU semantics and the close relationships between compound and basic expressions).  In this way, the issue of inferior transferable feature learning under large domain discrepancy and limited target samples can be greatly alleviated in CF-FER. 

% module explaination (totally update)
AUCH-Net consists of three modules: a backbone, an AU feature learning (AFL) module, and a visual feature learning (VFL) module. Specifically,  AUCH-Net first employs a backbone to extract basic features from the input. 
In the AFL module, the basic features are fed into an AU extraction block to extract a set of initial AU features. These features are modeled as nodes in the consistency-aware hypergraph, which is used to capture the relationships between AUs. 
 Meanwhile, in the VFL module, the consistency-aware hypergraph is also leveraged to capture the relationships between visual features and enhance feature representations. 
Finally, the features from the AFL and VFL modules are combined as the expression features. 
Notably, we take the original facial image and its flipped one as the input and develop a relation consistency loss to explicitly constrain the consistency between AUs on the hypergraphs. Moreover, we also propose an AU regularization loss to enable the learning of discriminative AU features. 
Thus, our method can 
learn consistent and discriminative feature representations in both the source and target domains, as illustrated in Fig.~\ref{fig:intro}(c). 

% contribution
Our contributions are given as follows:
\begin{itemize}
\item We develop a novel AUCH-Net for CF-FER by taking advantage of semantic-level AU features that are insensitive to large domain discrepancy and limited target samples. 
In this way, we can effectively obtain a highly transferable feature space for identifying compound expressions in the target domain with only a few samples.

\item We introduce an AFL module to learn consistent AU features. In particular, a relation consistency loss is developed to explicitly constrain the consistency between AUs on hypergraphs. By performing AU relation learning on consistency-aware hypergraphs, our method successfully models the intrinsic connections  between AUs and expression categories in the source domain.

\item We conduct extensive experiments on both in-the-lab and in-the-wild compound expression datasets to validate the superiority of our method over several state-of-the-art methods for CF-FER. Experimental results also clearly show the great potential of modeling AU relationships for CF-FER. 
\end{itemize}

The remainder of this paper is organized as follows. First, we 
	give the related work in Section \ref{relatedwork}. Then, we describe our proposed method in detail in Section \ref{method}.
	Next, we perform extensive experiments on the challenging compound expression datasets in Section \ref{experiments}. Finally, we draw the conclusion in Section \ref{conclusion}.

\section{Related Works}
\label{relatedwork}

In this section, we briefly review the related works closely related
to our method. %We first introduce basic FER and compound FER methods in Section \ref{basic}. Then, we introduce few-shot FER methods in Section \ref{few}.
%Next, we review hypergraph learning-based methods in Section \ref{hyper}. Finally, we introduce AU-assisted learning in Section \ref{au}. 

\subsection{Basic FER and Compound FER}
\label{basic}
% 新增
The basic FER task aims to classify a facial image into one of  basic expression categories.  
A variety of methods {\cite{ruan2021feature, zhang2024cf, li2025knowledge}} have been developed to extract expression features by performing disturbance disentanglement or attention learning 
for basic FER. For example, Du \emph{et al.}\cite{Du_Tao_Martinez_2014} reveal that facial images contain compound expressions, which enable a more subtle distinction of human emotional states. In practical applications, compound expressions play a crucial role in accurately capturing and interpreting human emotional states. Compared with the basic FER task, the compound FER task aims to identify compound expressions that exhibit more subtle differences.
Du  \emph{et al.}{~\cite{Du_Tao_Martinez_2014} collect facial images with compound expressions, enhancing the understanding of emotional states in practical applications.
% XXXX design XXXX for compound FER.
{Li  \emph{et al.}\cite{li2019separate} design a novel separate loss to maximize the intra-class similarity and minimize the inter-class similarity for compound FER. 
 Zhang \emph{et al.}\cite{kaminska2021two} propose a coarse-to-fine two-stage strategy to enhance the robustness of the learned compound expression features.} 

The aforementioned methods typically rely on a large number of high-quality annotated data for training. However, annotating compound expression data is not only time-consuming but also usually requires the use of professional psychological  knowledge to ensure data quality. Different from these methods, we study CF-FER, where the model is  only trained on basic expressions and evaluated on compound expressions. This largely alleviates the heavy burden of acquiring large-scale compound expression training data.

\subsection{Few-Shot FER}
\label{few}
Few-shot FER aims to identify novel expressions with very few samples based on the model learned from seen expressions. 
Ciubotaru\emph{et al.}~\cite{ciubotaru2019revisiting} 
formulate FER under the few-shot setting and revisit popular few-shot learning (FSL) methods in the cross-domain scenarios. %, where the base and novel classes belong to different domains. 
Shome \emph{et al.}~\cite{shome2021fedaffect} study 
few-shot FER under decentralized training. Zhu \emph{et al.}\cite{zhu2022convolutional} design a convolutional relation network, which imposes constraints to enhance feature representations. Note that the above methods are only concerned with  basic expressions. 

Later, Zou \emph{et al.}~\cite{zou2022facial}
study compound FER under the few-shot learning paradigm. %, which identifies novel compound expressions with the model learned from basic expressions. 
Instead of dividing a compound expression dataset into a base class set and a novel class set, they study 
more practical CF-FER, where the base class set and the novel class set are 
 easily accessible basic expression datasets and the compound expression dataset, respectively. Accordingly, they develop a two-branch EGS-Net to transfer the knowledge from the source domain to the target domain. % by performing joint and alternate learning between the two branches. 
Zou \emph{et al.}~\cite{zou2022learn}
propose a novel CDNet, which is trained by cascading parameter-shared decomposition modules, for CF-FER. Chen \emph{et al.}~\cite{chen2025hyperbolic} propose a hyperbolic self-paced multi-expert network (HSM-Net) for CF-FER. HSM-Net uses hyperbolic space-based mixture-of-experts layers to alleviate imbalanced expression categories and a hyperbolic self-paced learning strategy to model hierarchical expression relationships, enhancing knowledge transfer from basic to compound expressions.
Due to large domain discrepancy and limited target samples, the performance of existing CF-FER methods is still inferior. 

In this paper, we follow the same  setup as~\cite{zou2022learn,zou2022facial,chen2025hyperbolic}. 
In contrast to existing methods, we develop an AUCH-Net by performing consistency-aware hypergraph learning on AUs, facilitating 
learning consistent feature representations.

\subsection{Hypergraph Learning}
\label{hyper}
{Ruan \emph{et al.}\cite{ruan2021feature}} introduce graph neural networks (GNNs) to learn inter-relationships between features. 
% However, GNNs can only model pairwise relations between nodes. 
{
However, GNNs only construct edges between two nodes, limiting their modeling capability.}
Different from GNNs, hypergraph neural networks (HGNNs) {\cite{zhou2006learning}} can encode high-order feature correlations, capturing different levels of relational structures. 
 Jiang \emph{et al.}~\cite{jiang2019dynamic} develop the  dynamic HGNN, which uses $K$-means and $K$ nearest neighbors (KNN) to dynamically construct hypergraphs.
Han \emph{et al.}~\cite{han2023vision}  introduce a vision HGNN, which alternately
performs patch embedding and hypergraph construction to enhance structure-aware image representations for image classification.

Unlike existing methods, we construct the consistency-aware hypergraph on AUs to capture high-order relationships between AUs by introducing a novel relation consistency loss. Hence, the relationships between AUs are well exploited and adapted to infer new compound expressions.

\subsection{AU-Assisted Learning}
\label{au}
Action units (AUs) are defined to describe the movement and activation states of key facial muscles, and they have been widely applied in facial feature localization and emotion understanding \cite{mao2025facial, belharbi2024guided}.
Peng \emph{et al.}~\cite{peng2019dual} exploit the relationships between AUs encoding and expression categories to automatically annotate AUs for facial images with only expression labels. 
Niu \emph{et al.}~\cite{niu2019multi} propose two networks to generate multi-view features for both labeled and unlabeled facial images, and use a multi-label co-regularization loss to minimize the distance of the predicted AU probability distributions from the two views. 
Lan \emph{et al.}~\cite{lan2024expllm} employ a large-scale language model to analyze AU interactions, identifying dominant sentiments and generating accurate chains of thought for FER.
Belharbi \emph{et al.}~\cite{belharbi2024guided} leverage prior relationships between AUs and expressions to generate attention maps, obtaining interpretable feature representations.

Different from existing methods, we propose to perform hypergraph learning on AUs, 
enabling our method to effectively learn the intrinsic connections between AUs and expression categories. 
Such a way greatly 
 addresses the problem of inferior transferable feature learning under large domain discrepancy and limited target samples in CF-FER.

\section{Proposed Method}
\label{method}
{In this section, we describe our proposed method in detail. 
	First, we give the preliminaries and notations in Section \ref{problem}. Then, we provide the overview of our method in Section \ref{overview}. Finally, we present technical details of the AU feature learning module and the visual feature learning module in Section \ref{ldd} and Section \ref{gdd}, respectively. 
    
\subsection{Preliminaries and Notations}
\label{problem}
%\noindent %\textbf{Settings} 
Following \cite{zou2022learn,zou2022facial}, we perform compound FER under the cross-domain FSL setting, where the model is trained on the base class set (i.e., multiple basic expression datasets in the source domain) and evaluated on the novel class set (i.e., a compound expression dataset in the target domain). 

\noindent \textbf{Notations} The base class set is denoted as $\mathcal{D}_{b} = \{(\textbf{X}^{b},y^{b})\}$, where 
$\textbf{X}^{b}$ denotes the basic expression image in $\mathcal{D}_{b}$ and $y^{b}$ denotes the corresponding category label. The novel class set is denoted as  
$\mathcal{D}_{c} = \{(\textbf{X}^{c},y^{c})\}$, where 
%$\textbf{X}^{c}$ denotes the compound expression image in $\mathcal{D}_{c}$ and $y^{c}$ denotes the corresponding category label. 
$\mathcal{D}_{base}$ contains only the basic expressions while $\mathcal{D}_{novel}$ contains compound expressions.
Note that the base class set and the novel class set do not overlap. ($\mathcal{C}_{base} \cap \mathcal{C}_{novel} = \emptyset$).

During the training phase, given an input image $\textbf{X}^b$ in the base class set, we get its flipped counterpart $\textbf{X}_{flip}^b$.
The two images are taken as the input to the backbone for extracting basic features  %We denote these features as 
$\textbf{F} \in \mathbb{R}^{C \times W \times H}$ and $\textbf{F}_{flip} \in \mathbb{R}^{C \times W \times H}$. %, where $\textbf{F}$ and $\textbf{F}_{flip}$ denote the basic features corresponding to $\textbf{X}^b$ and $\textbf{X}_{flip}^b$, respectively. 
Here $C$, $W$, and $H$ denote the channel number, the width, and the height of the basic feature, respectively. During the testing phase, given a support set $\mathcal{S}$ (containing $N$ classes with $K$ images per class) and a query set $\mathcal{Q}$, the trained model classifies each image in the query set into one of  the categories in the support set.

\subsection{Overview}
\label{overview}
Fig.~\ref{fig: main} gives the overview of our AUCH-Net, which consists of three modules: a backbone, an AU feature learning (AFL) module, and a visual feature learning (VFL) module. 
The AFL module first extracts initial AU features by an AU extraction block. Then, it constructs two consistency-aware hypergraphs (corresponding to  $\textbf{X}^b$ and $\textbf{X}^b_{flip}$) via a parameter-shared 
HGNN 
to 
capture the relationships between AUs, where a relation consistency loss is introduced to explicitly impose constraints on the consistency between AU features (extracted from $\textbf{X}^b$ and $\textbf{X}^b_{flip}$).  
In addition, an AU regularization loss is developed to regularize the learning of AU features, facilitating obtaining discriminative AU features. The VFL module models high-order relationships between basic features via another two consistency-aware hypergraphs. The relation consistency loss and the classification loss are employed to learn visual features. By combining AU and visual features, a  transferable expression feature is finally obtained for a facial image.

\begin{figure*}[ht!]
    \centering
    \includegraphics[width=0.99\textwidth]{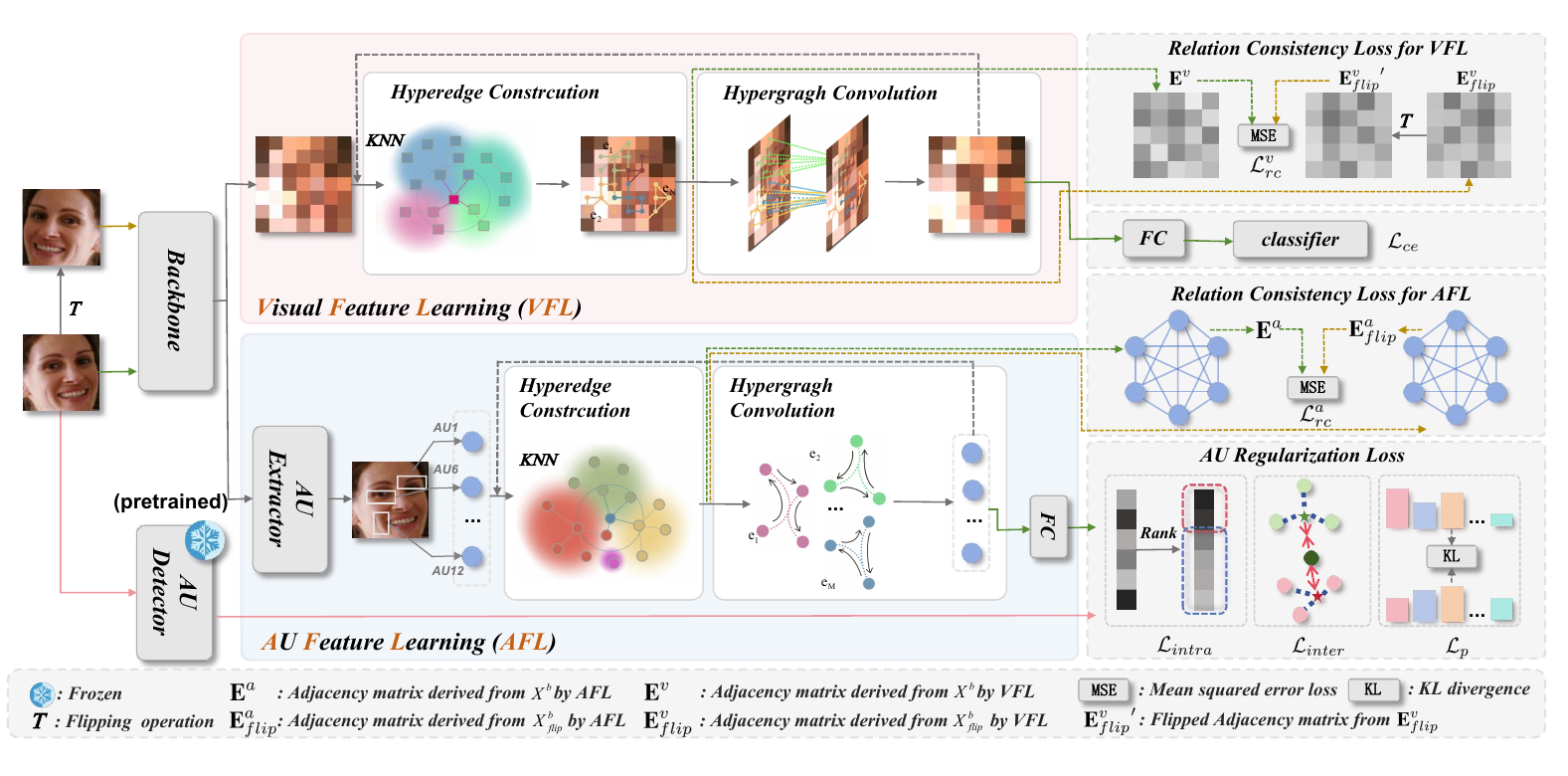}
    \caption{Overview of our AUCH-Net. AUCH-Net consists of a backbone, an AU feature learning (AFL) module, and a visual feature learning 
(VFL) module. The AFL module extracts AU features under the guidance of  a relation consistency loss and an AU regularization loss, while the VFL module extracts visual features under the guidance of a relation consistency loss and a classification loss. }
    \label{fig: main} 
\end{figure*}

\subsection{AU Feature Learning (AFL) Module}
\label{ldd}
The basic features $\textbf{F}$ and $\textbf{F}_{flip}$ are first fed into an AU extraction block (consisting of a convolutional layer, a reshaping layer, and an average pooling layer) to extract two sets of initial AU features (i.e., $\textbf{A}^{0} = [\textbf{a}_1^{0}, \dots, \textbf{a}_M^{0}] \in \mathbb{R}^{{D \times M}}$ and $\textbf{A}_{flip}^{0} = [\textbf{a}_{flip,1}^{0}, \dots, \textbf{a}_{flip,M}^{0}] \in \mathbb{R}^{{D \times M}}$. Here, $M$ denotes the number of initial AU features and $D$ is the feature dimension). Based on $\textbf{A}^{0}$ and $\textbf{A}^{0}_{flip}$, two consistency-aware hypergraphs are constructed and updated via a parameter-shared HGNN to capture the relationships between AUs. 
A relation consistency loss is 
introduced to explicitly learn consistent AU features. In this way, 
 two sets of consistency-aware AU features $\textbf{U}^a = {[\textbf{u}^a_1, \dots, \textbf{u}^a_M]} \in \mathbb{R}^{{D \times M}}$ and $\textbf{U}_{flip}^a = {[\textbf{u}^a_{flip,1}, \dots,\textbf{u}^a_{flip,M}]} \in \mathbb{R}^{{D \times M}}$ corresponding to $\textbf{X}^b$ and $\textbf{X}_{flip}^b$ can be obtained. 
To learn discriminative AU features, we introduce an AU regularization loss, which imposes proper regularization on AU learning. Finally, a fully-connected (FC) layer is used to predict AU scores.

\subsubsection{Consistency-Aware Hypergraph}
\label{section: consistency aware hypergragh}
Given $\textbf{A}^{0}$, we introduce a consistency-aware hypergraph via an HGNN \cite{zhou2006learning,jiang2019dynamic} to model the relationships between AUs. 
The consistency-aware hypergraph is defined as $\textbf{G}_s = (\nu_s, \varepsilon_s, \textbf{W}_s)$, where $\nu_s$ denotes the node set, $\varepsilon_s$ denotes the hyperedge set, and $\textbf{W}_s$ denotes the weight matrix of the hyperedge set. 
% Here, {$M$} denotes the number of nodes and ${K_s}$ denotes the number of hyperedges. 
We use a set of initial AU features %$\textbf{a}_i$ $(i=1, \dots, M)$
as nodes and apply the $K$-nearest neighbors (KNN) algorithm to construct the hyperedges. 
Generally, the construction of the hypergraph relies on the AU features. Meanwhile, the AU features are learned and updated by hypergraph convolution. 
To fully capture the relationships between AUs, we employ alternate learning between hypergraph construction and AU feature learning. Such a way ensures
mutual improvements of meaningful AU features and
an effective hypergraph structure.

Specifically, at the $l$-th iteration, the AU features $\textbf{A}^{(l-1)}$ learned at the $(l-1)$-th iteration are first taken as the nodes for hyperedge construction. We construct the hyperedge based on the KNN algorithm, that is, 
\begin{equation}
  \textbf{H}^{l} = \mathrm{F}(\mathbf{A}^{(l-1)}, \epsilon),
  \label{eq:l_hl}
\end{equation}
where $\textbf{H}^{l}$ denotes the incidence matrix; $\epsilon$ is the threshold to filter out neighbors with large distances; $\mathrm{F}(\cdot,\cdot)$ denotes the hyperedge construction operation which identifies  $K$ nearest neighbors of each node with the Euclidean distance.

Then, we apply a  hypergraph convolutional layer to aggregate high-order structure information and enhance feature representations. The AU features are updated by
\begin{equation}
  \textbf{A}^{l} = \textbf{E}^{l} \textbf{A}^{(l-1)} \Theta,
  \label{eq:l_al}
\end{equation}
where {$\textbf{E}^{l} = \textbf{D}^{-1/2}_v \textbf{H}^l \textbf{W}_s \textbf{D}^{-1}_e (\textbf{H}^l)^\mathrm{T} \textbf{D}^{-1/2}_v$} denotes the adjacency matrix; $\textbf{D}_v$ and $\textbf{D}_e$ represent the diagonal matrices of vertex degrees and edge degrees, respectively; $\Theta$ denotes the learnable parameters.

After $L$ iterations, we obtain a set of AU features $\textbf{U}^a$ and the incidence matrix $\textbf{H}^a \in \mathbb{R}^{M \times M}$, and the adjacency matrix $\textbf{E}^{a} \in \mathbb{R}^{M \times M}$ for the input image $
\textbf{X}^b$. Analogously, given $\textbf{A}^0_{flip}$, we construct 
a consistency-aware hypergraph and apply hypergraph convolution (we use parameter-shared convolutional layers for $\textbf{A}^0$ and $\textbf{A}^0_{flip}$), obtaining a set of AU features $\textbf{U}^a_{flip}$, the incidence matrix $\textbf{H}^a_{flip}$, and the adjacency matrix $\textbf{E}^a_{flip}$ for $\textbf{X}_{flip}^b$.

Finally, we use an FC layer to predict AU scores, i.e.,  
\begin{equation}
  \textbf{v} = \sigma(\textbf{W}_c^\mathrm{T} \textbf{U}^a),
  \label{eq:V_pred}
\end{equation}
where $\textbf{W}_c \in \mathbb{R}^{D
\times 1}$ denotes the parameters of the FC layer; $\sigma$ is the ReLU activation function;
$\textbf{v} \in \mathbb{R}^{M\times 1}$ denotes the AU scores. Each element in $\textbf{v}$ represents the probability of the occurrence for each AU. 

\noindent \textbf{Relation Consistency Loss}
To ensure that the model can learn consistent feature representations under different spatial transformations of the image, inspired by \cite{guo2019visual, zhang2022learn},
we preserve the relationships between AU features extracted from the original image and its flipped version.
%Hence, we introduce a relation consistency loss.
As mentioned previously, two adjacency matrices, which denote the similarities between AUs, are respectively learned for $\textbf{X}^b$ and $\textbf{X}^b_{flip}$. 
Hence, the relation consistency loss is defined as
\begin{equation}
  \mathcal{L}_{rc}^a = \frac{1}{M^2} \sum_{i=1}^M\sum_{j=1}^M (\textbf{E}^a(i,j) - \textbf{E}^a_{flip}(i,j))^2,
  \label{eq:l_rc_1}
\end{equation}
where 
$\textbf{E}^a(i,j)$ and $\textbf{E}^a_{flip}(i,j)$ denote the elements in the $i$-th row and $j$-th column of $\textbf{E}^a$ and $\textbf{E}^a_{flip}$, respectively.

Instead of imposing attention consistency on raw image patches  \cite{guo2019visual, zhang2022learn}, we perform relation consistency learning on hypergraphs to learn consistent relations between semantic-level AU features, greatly facilitating CF-FER.

\subsubsection{AU Regularization Loss} 
To learn discriminative AU features, we propose an AU regularization loss, which contains three losses: 1) an \textit{intra-AU loss} that enhances the differentiation between different AU scores; 2) an \textit{inter-AU loss} that reduces the distances of AU features from the same category while enlarging the distances of AU features from the different categories; 3) a \textit{prior loss} that enables the learning of appropriate AU features under the guidance of a pretrained AU detector.

\noindent \textbf{Intra-AU Loss}
Each expression is associated with several activated AUs instead of all AUs (see Fig.~\ref{fig:intro}). In this paper, we employ the intra-AU loss (same as the rank regularization loss  \cite{wang2020suppressing}) by sorting the scores corresponding to all AUs and explicitly imposing a margin on the differences between high scores (corresponding to activated AUs) and low scores (corresponding to non-activated AUs).
% \begin{comment}
% Specifically, we first sort the scores corresponding to all AUs 
% in ascending order, and then we divide these scores into two groups (one with lower scores and the other with higher scores) according to a threshold $\gamma$. Next, we calculate the average scores of the two groups separately and impose a margin to discriminate the 
% two average scores. 
% \end{comment}
Mathematically, the intra-AU loss is defined as
\begin{equation}
  \mathcal{L}_{intra} = \textrm{max}\{0, \delta - (\alpha_H -\alpha_L) \},
  \label{eq:l_rr}
\end{equation}
where $\delta$ represents the margin; $\alpha_H$ denotes the average score of the group with high scores; $\alpha_L$ denotes the average score of the group with low scores.

\noindent \textbf{Inter-AU loss}
When two facial images correspond to the same expression category, their predicted AU scores should be close. When two facial images belong to different expression categories, their predicted AU scores should be far away. Motivated by the center loss \cite{wen2016discriminative}, we develop the inter-AU loss. Technically, 
we first obtain the AU score center of each category 
and compute the distances between the input AU scores and the centers. Based on this, 
we enhance the intra-class compactness and inter-class separability. Hence, 
the inter-AU loss is defined as
\begin{equation}
  \mathcal{L}_{inter} = \sum_i {{\mathbb{I}_{[y=i]}}\textrm{log}(1+d_i) + (1-{\mathbb{I}_{[y=i]}})e^{-d_i}},
  \label{eq:l_cmp}
\end{equation}
where $\mathbb{I}_{[y=i]}$ equals 1 when $y=i$, and 0 otherwise;
$y \in [1,C_{base}]$ denotes the category label of the input image $\textbf{X}^b$; $C_{base}$ denotes the number of basic expressions;
$d_i$ denotes the distances (we use the Kullback-Leibler (KL) divergence) between the input AU scores and the AU score center $\textbf{c}_i$ of the $i$-th expression category, that is, 
 $ d_i = \sum_{j=0}^M \textbf{v}(j) \log [\textbf{v}(j)/{\textbf{c}_i(j)}]$;
$\textbf{c}_i(j)$ is the $j$-th element in the center $\textbf{c}_i$. The inter-AU loss consists of two terms. The first term ensures the compactness of the AU scores $\mathbf{v}$ from the same category, while the second term ensures the differences of AU scores from the different categories. Note that different from {the contrastive loss \cite{hadsell2006dimensionality} and the triplet loss \cite{schroff2015facenet}},
we use the KL divergence to faithfully measure the distance between AU scores. Moreover, we introduce the exponential function (instead of the logarithm) to ensure that the loss value remains greater than 0 and avoids generating excessively large values.

\noindent \textbf{Prior Loss}
To learn appropriate AU scores, we employ a pretrained AU detector with good generalization performance  \cite{yin2024fg} to generate soft labels for each expression image:
\begin{equation}
  \textbf{v}_{soft} = \mathrm{AD}(\textbf{X}^b),
  \label{eq:y_soft}
\end{equation}
where $\mathrm{AD}(\cdot)$  denotes the pretrained AU detector and $\textbf{v}_{soft} \in \mathbb{R}^{M \times 1}$ denotes the generated soft label.

 Then, we compute the distance  between the predicted AU scores and $\textbf{v}_{soft}$ by the KL divergence:
\begin{equation}
  \mathcal{L}_{p} = \sum_{i=0}^M \textbf{v}(i) \log \frac{\textbf{v}(i)}{\textbf{v}_{soft}(i)}.
  \label{eq:l_guide}
\end{equation}

Based on the above, the AU regularization loss is
\begin{equation}
  \mathcal{L}_{au} = \mathcal{L}_{intra} + \mathcal{L}_{inter} + \mathcal{L}_{p}.
  \label{eq:l_regu}
\end{equation}

\subsection{Visual Feature Learning (VFL) Module}
\label{gdd}
The basic features $\textbf{F}$ and $\textbf{F}_{flip}$ are fed into a reshaping layer to obtain two sets of initial visual features $\textbf{P}^0 = [\textbf{p}^0_1, \dots, \textbf{p}^0_N ] \in \mathbb{R}^{{C \times N}}$ and $\textbf{P}^0_{flip} = [\textbf{p}^{0}_{flip,1}, \dots, \textbf{p}^{0}_{flip,N}] \in \mathbb{R}^{{C \times N}}$. Here $N = H \times W$ denotes the number of visual features, and $C$ denotes the dimension of each feature. 
In the AFL module, we construct the relationship between AU features. In the VFL module, we  construct two consistency-aware hypergraphs to learn the relationships between pixels. 
Given $\textbf{P}^0$, 
the consistency-aware hypergraph 
is defined as $\textbf{G}_v = (\nu_v, \varepsilon_v, \textbf{W}_v)$, where $\nu_v$ denotes the node set, $\varepsilon_v$ denotes the hyperedge set, and 
$\textbf{W}_v$ denotes the weight matrix. 
We use a set of initial visual features as nodes and use the KNN algorithm to construct the hypergraph. 
Similar to hypergraph construction and updating in the previous section, after $L$ iterations, we obtain the visual features $\textbf{U}^{v}{ = [\textbf{u}^v_1, \dots, \textbf{u}^v_N ] \in \mathbb{R}^{{C \times N}}}$ and the incidence matrix $\textbf{H}^{v} \in \mathbb{R}^{{N \times N}}$, and the adjacency matrix $\textbf{E}^{v} { \in \mathbb{R}^{N \times N}}$ for the input image.
Analogously, given $\textbf{P}^0_{flip}$, we can obtain $\textbf{U}^{v}_{flip}{ = [\textbf{u}^v_{flip,1}, \dots, \textbf{u}^v_{flip,N} ] \in \mathbb{R}^{{C \times N}}}$   and the incidence matrix $\textbf{H}^{v}_{flip} \in \mathbb{R}^{{N \times N}}$ and the adjacency matrix $\textbf{E}^{v}_{flip} { \in \mathbb{R}^{N \times N}}$.
The relation consistency loss is given as
\begin{equation}
  \mathcal{L}_{rc}^v  = \frac{1}{{N^2}} \sum_{i=1}^N\sum_{j=1}^N (\textbf{E}^v(i,j) - {\textbf{E}^v_{flip}}'(i,j))^2,
  \label{eq:l_rc_2}
\end{equation}
where ${\textbf{E}^v_{flip}}' = \mathrm{FLIP}({\textbf{E}^v_{flip}})$ and $\mathrm{FLIP}(\cdot)$ denotes a mapping function to align ${\textbf{E}^v_{flip}}$ with ${\textbf{E}^v}$ in positions.

Finally, we apply an FC layer to predict the expression category and calculate the classification loss as
\begin{equation}
  \mathcal{L}_{ce} = - \sum_{i=1}^{{C}_{base}} \mathbb{I}_{[i=y]}\mathrm{log}(\mathrm{F_c}(\textbf{U}^{v})),
  \label{eq:l_ce}
\end{equation}
where $\mathrm{F_c}(\cdot)$ is an FC layer and $\mathbb{I}_{[i=y]}$ equals 1 when $i=y$, and 0 otherwise. %$y_{e}$ denotes the label of $\textbf{X}_b$.

Based on the above, the overall loss is
\begin{equation}
  \mathcal{L} = \mathcal{L}_{ce} + \lambda_{rc}^v \mathcal{L}_{rc}^v + \lambda_{rc}^a \mathcal{L}_{rc}^a + \lambda_{au} \mathcal{L}_{au},
  \label{eq:l_sum}
\end{equation}
where $\lambda_{rc}^v$, $\lambda_{rc}^a$, and $\lambda_{au}$ denote the loss weights.

\section{Experiments}
\label{experiments}
In this section, we first introduce the experimental settings in Section \ref{datasets}. Then, we present the implementation details in Section \ref{implementation}. Next, we conduct ablation studies in Section \ref{abltion}. After that, we give some visualization results and computational complexity in Section \ref{visualization} and Section \ref{computation}, respectively. Finally, we compare our method with several state-of-the-art methods in Section \ref{SOTA}.

%\subsection{Experimental Settings}

\subsection{Datasets}
\label{datasets}
Our AUCH-Net is trained on multiple basic expression datasets and tested on the compound expression dataset. In this paper, we use five basic expression datasets (including three in-the-lab datasets (CK+ \cite{lucey2010extended}, MMI \cite{Pantic_Valstar_Rademaker_Maat_2005}, and Oulu-CASIA \cite{Zhao_Huang_Taini_Li_Pietikäinen_2011}), and two in-the-wild datasets (RAF-DB \cite{8099760} and SFEW \cite{dhall2011static})) to construct the training set. Three compound expression datasets (CFEE\_C \cite{Du_Tao_Martinez_2014}, EmotioNet\_C \cite{Benitez-Quiroz_Srinivasan_Martinez_2016}, and RAF\_C \cite{8099760}) are used to evaluate the performance of the learned model.

\noindent \textit{Basic Expression Datasets.} CK+ contains 593 video sequences, where 327 video sequences are annotated with seven basic expressions. MMI contains 326 video sequences with six basic expressions. % (we use 205 frontal-view sequences). 
Oulu-CASIA consists of 2,880 video sequences with six basic expressions. % (we use 480 indoor video sequences). 
Three peak frames of each sequence in the above in-the-lab datasets are selected for training. RAF-DB consists of a basic subset (involving 12,271 training images) with seven basic expressions. SFEW is labeled with seven basic expressions with 958 training images. All the samples in RAF-DB and SFEW are used for training. %Note that the imbalance ratio of the training set is up to 1:114 (the ratio between the contempt expressions and the happy expressions), indicating significantly imbalanced expression categfigureories.

\noindent \textit{Compound Expression Datasets.} CFEE\_C is derived from the CFEE dataset. It is an in-the-lab dataset and annotated with 15 compound expressions from 230 subjects, including a total of 5,046 facial images. EmotioNet\_C is collected from the EmotioNet challenge, where the samples are collected in the wild and annotated with 10 different compound expressions. It contains 2,471 facial images.
RAF\_C is the compound subset of RAF-DB with 11 compound expressions and a total of 3,162 facial images.

\begin{comment}
\noindent \textit{Basic Expression Datasets.} CK+ contains 593 video sequences, where 327 video sequences are annotated with seven basic expressions. MMI contains 326 video sequences with six basic expressions. % (we use 205 frontal-view sequences). 
Oulu-CASIA consists of 2,880 video sequences with six basic expressions. % (we use 480 indoor video sequences). 
Three peak frames of each sequence in the above in-the-lab datasets are selected for training. RAF-DB consists of a basic subset (involving 12,271 training images) with seven basic expressions. SFEW is labeled with seven basic expressions with 958 training images. All the samples in RAF-DB and SFEW are used for training. %Note that the imbalance ratio of the training set is up to 1:114 (the ratio between the contempt expressions and the happy expressions), indicating significantly imbalanced expression categories.

\noindent \textit{Compound Expression Datasets.} CFEE\_C is derived from the CFEE dataset. It is an in-the-lab dataset and annotated with 15 compound expressions from 230 subjects, including a total of 5,046 facial images. EmotioNet\_C is collected from the EmotioNet challenge, where the samples are collected in the wild and annotated with 10 compound expressions. It contains 2,471 facial images.
RAF\_C is the compound subset of RAF-DB with 11 compound expressions and a total of 3,162 facial images.
\end{comment}

\subsection{Implementation Details}
\label{implementation}

Our method is implemented by Pytorch. We employ RFS \cite{tian2020rethinking} as our Baseline method, where ResNet-18 \cite{He_Zhang_Ren_Sun_2016} pretrained on MS-Celeb-1M \cite{Guo_Zhang_Hu_He_Gao_2016} is used as the backbone. All the facial images are first aligned and resized to the size of $256 \times 256$. Then, they are randomly cropped to the size of $224 \times 224$, followed by a random horizontal flip and color jitter as data augmentation. We use FG-Net\cite{yin2024fg} pretrained on the BP4D\cite{zhang2014bp4d} dataset as our AU detector. Following \cite{zou2022learn}, we use a mapping function to unify the expression labels in basic expression datasets. Our method is trained using the stochastic gradient descent (SGD) optimizer with a learning rate of $0.1$ and a weight decay of $5 \times {10}^{-4}$.  For the training stage, the model is trained for $80$ epochs. 
The margin $\delta$ in  Eq. (\ref{eq:l_rr}) is set to 0.2 and the threshold $\gamma$ to divide AU scores is set to 0.25. For hyperedge construction, the value $\epsilon$ in Eq. (\ref{eq:l_hl}) is set to 0.5.
The value $\lambda_{rc}^v$, $\lambda_{rc}^a$, and $\lambda_{au}$ in Eq. (\ref{eq:l_sum}) are set to 0.5, 0.5, and 0.1, respectively.
For a few-shot task, we set the number of classes $N$ = 5, the number of support samples $K$ = 1 or 5, and the number of query samples $Q$ = 16 for each class.
{Test accuracy (\%) of 5-way few-shot classification tasks with 95\% confidence intervals is reported as the evaluation metric. 
%Similar to representative FSL methods\cite{snell2017prototypical}, we assign the query images to their nearest classes in the learned expression feature space.}

For training, we train our model on the base class set using batch training. For testing, an  $N$-way  $K$-shot support set is randomly sampled from the novel class set. Given an input image, we obtain the visual features and AU features, and concatenate them as an expression feature. Similar to representative FSL methods \cite{snell2017prototypical}, we assign the query images to their nearest classes in the learned expression feature space.

% and then calculate the cosine distance between this feature and class prototypes.

We conduct our experiments on an NVIDIA GeForce RTX 3090 GPU. The training is performed using a batch training manner. For data pre-processing, all facial images are first aligned and resized to $256 \times 256$. Subsequently, they are randomly cropped to $224 \times 224$. To enhance the robustness of our model, we apply data augmentation techniques, including random horizontal flipping and color jitter. In addition, we employ random erasing %with a scale parameter set to $(0.02, 0.25)$, which further enriched 
to enrich the training data by randomly erasing parts of the images. During the testing phase, we use the outputs from both the AFL module and the VFL module to fine-tune the classifier. To ensure reproducibility, we set the random seed to 3407 in all our experiments.

\subsection{Ablation Studies}
\label{abltion}

\begin{comment}
\begin{table}[!t]
\footnotesize
     \caption{{The details of seven variants of our method.}}
    \renewcommand\arraystretch{0.7}{
    \resizebox{\linewidth}{!}{
	\setlength{\tabcolsep}{1.89mm}{
		\begin{tabular}{l c c c c c }
			%\begin{tabular}{l|c|c|c|c}
			%\toprule [2 pt]
                  %\toprule
        \toprule[0.6pt]
        \tiny{Method} & \tiny{AFL} &\tiny{VFL} & \tiny{$\mathcal{L}_{rc}^a$} & \tiny{$\mathcal{L}_{rc}^v$} & \tiny{$\mathcal{L}_{au}$}\\
		  \midrule[0.4pt]
	      \tiny{Baseline}	        & \ding{55} & \ding{55} & \ding{55}  & \ding{55}      & \ding{55} \\
        \tiny{AUCH-Net\_{au}}     & \ding{55} & \ding{55} & \ding{55}  & \ding{55}      & \ding{52} \\
        \tiny{AUCH-Net\_{a/r}}    & \ding{52}  & \ding{55} & \ding{55}  & \ding{55}      & \ding{52} \\
        \tiny{AUCH-Net\_a}        & \ding{52}  & \ding{55} & \ding{52}   & \ding{55}      & \ding{52}  \\
        \tiny{AUCH-Net\_{v/r}}    & \ding{55} & \ding{52}  & \ding{55}  & \ding{55}      & \ding{55} \\
        \tiny{AUCH-Net\_v}	      & \ding{55} & \ding{52}  & \ding{55}  & \ding{52}       & \ding{55} \\
        \tiny{AUCH-Net}           & \ding{52}  & \ding{52}  & \ding{52}   & \ding{52}       & \ding{52} \\
		\bottomrule[0.6pt]
                %\bottomrule
		\end{tabular}
	}}
 }

 \label{tab:variants of aurynet}
\end{table}
\end{comment}

\begin{table}[!t]
\footnotesize
    \caption{{{The details of seven variants} of our proposed method. `AFL' denotes the the AFL module while `VFL' denotes the VFL module.}}
    \renewcommand\arraystretch{1.0}{
    \resizebox{\linewidth}{!}{
	\setlength{\tabcolsep}{1.89mm}{
		\begin{tabular}{l| c c c c c }
			%\begin{tabular}{l|c|c|c|c}
			%\toprule [2 pt]
                  %\toprule
		\hline
        Method & AFL &VFL & $\mathcal{L}_{rc}^a$ & $\mathcal{L}_{rc}^v$ & $\mathcal{L}_{au}$\\
		 \hline
	Baseline	      & $\times$ & $\times$ & $\times$ & $\times$   & $\times$ \\
        AUCH-Net\_{au}     & $\times$ & $\times$ & $\times$ & $\times$   & $\surd$ \\
        AUCH-Net\_{a/r}   & $\surd$ & $\times$ & $\times$ & $\times$    & $\surd$ \\
        AUCH-Net\_a       & $\surd$ & $\times$ & $\surd$ & $\times$    & $\surd$  \\
        AUCH-Net\_{v/r}   & $\times$ &$\surd$   & $\times$ & $\times$   & $\times$ \\
        AUCH-Net\_v	      & $\times$ & $\surd$  & $\times$  & $\surd$   & $\times$ \\
        AUCH-Net          & $\surd$ & $\surd$  & $\surd$  & $\surd$    & $\surd$ \\
		\hline
                %\bottomrule
		\end{tabular}
	}}
 }
 \label{tab:variants of aurynet}
\end{table}

We perform ablation studies on seven variants of AUCH-Net  on CFEE\_C and RAF\_C. 
The details of seven variants are given in Table \ref{tab:variants of aurynet}. The abaltion study results are shown in Table \ref{tab:ablation of AUCH-Net}. %More results are given in the Supplement. 

\begin{table}[!t]
\footnotesize
         \caption{{Ablation study results obtained by seven variants of our AUCH-Net. The test accuracy (\%) of 5-way few-shot classification tasks with 95\% confidence intervals is reported. }}
    \renewcommand\arraystretch{1.1}{
    \resizebox{\linewidth}{!}{
	\setlength{\tabcolsep}{1.89mm}{
		\begin{tabular}{l| c c | c c }
			%\begin{tabular}{l|c|c|c|c}
			%\toprule [2 pt]
                  %\toprule
		\hline
        \multirow{2}{*}{Method} & \multicolumn{2}{c|}{CFEE\_C} & \multicolumn{2}{c}{RAF\_C} \\
	   & 1-shot & 5-shot & 1-shot  & 5-shot \\ \hline
Baseline	   & 55.21 \tiny{$\pm$ 0.71} & 66.45 \tiny{$\pm$ 0.43} & 43.31 \tiny{$\pm$ 0.63} & 61.11 \tiny{$\pm$ 0.53} \\
AUCH-Net\_{au}     & 56.77 \tiny{$\pm$ 0.64} & 68.34 \tiny{$\pm$ 0.58} & 46.82 \tiny{$\pm$ 0.70} & 63.01 \tiny{$\pm$ 0.55} \\
AUCH-Net\_{a/r} & 57.21 \tiny{$\pm$ 0.48} & 68.70 \tiny{$\pm$ 0.73} & 47.17 \tiny{$\pm$ 0.67} & 63.23 \tiny{$\pm$ 0.49} \\
AUCH-Net\_a     & 57.79 \tiny{$\pm$ 0.43} & 69.10 \tiny{$\pm$ 0.44} & 47.98 \tiny{$\pm$ 0.45} & 63.70 \tiny{$\pm$ 0.56} \\
AUCH-Net\_{v/r} & 56.09 \tiny{$\pm$ 0.78} & 68.12 \tiny{$\pm$ 0.70} & 45.63 \tiny{$\pm$ 0.57} & 61.98 \tiny{$\pm$ 0.67} \\
AUCH-Net\_v	   & 56.88 \tiny{$\pm$ 0.73} & 68.30 \tiny{$\pm$ 0.44} & 46.66 \tiny{$\pm$ 0.64} & 62.38 \tiny{$\pm$ 0.74} \\
AUCH-Net        & \textbf{58.36} \tiny{$\pm$ 0.92} & \textbf{70.11} \tiny{$\pm$ 0.74} & \textbf{48.65} \tiny{$\pm$ 0.68} & \textbf{64.55} \tiny{$\pm$ 0.49} \\
		\hline
                %\bottomrule
		\end{tabular}
	}}
 }
 \label{tab:ablation of AUCH-Net}
\end{table}

\begin{comment}
\begin{table}[!t]
\footnotesize
     \caption{{Ablation study results obtained by seven variants of our AUCH-Net. The test accuracy (\%) of 5-way few-shot classification tasks with 95\% confidence intervals is reported. }}
    \renewcommand\arraystretch{1.05}{
    \resizebox{\linewidth}{!}{
	\setlength{\tabcolsep}{1.89mm}{
		\begin{tabular}{l c c  c c }
			%\begin{tabular}{l|c|c|c|c}
			%\toprule [2 pt]
                  %\toprule
		% \hline
        \toprule[1.0pt]
        \multirow{2}{*}{Method} & \multicolumn{2}{c}{CFEE\_C} & \multicolumn{2}{c}{RAF\_C} \\
	   & 1-shot & 5-shot & 1-shot  & 5-shot \\ 
       \midrule[0.8pt]
Baseline	       & 55.21 & 66.45  & 43.31 & 61.11 \\
AUCH-Net\_{au}     & 56.77 & 68.34  & 46.82 & 63.01 \\
AUCH-Net\_{a/r}    & 57.21 & 68.70  & 47.17 & 63.23 \\
AUCH-Net\_a        & 57.79 & 69.10  & 47.98 & 63.70 \\
AUCH-Net\_{v/r}    & 56.09 & 68.12  & 45.63 & 61.98 \\
AUCH-Net\_v	       & 56.88 & 68.30  & 46.66 & 62.38 \\
AUCH-Net           & \textbf{58.36} & \textbf{70.11} & \textbf{48.65} & \textbf{63.90} \\
		% \hline
        \bottomrule[1.0pt]
		\end{tabular}
	}}
 }

 \label{tab:ablation of AUCH-Net}
\end{table}
\end{comment}

\noindent \textbf{Effectiveness of Hypergraph Modeling}
We employ HGNN to construct high-order relationships in both  AFL and VFL modules. As shown in Table \ref{tab:ablation of AUCH-Net}, 
 AUCH-Net\_{a/r} outperforms AUCH-Net\_{au}. 
Meanwhile, 
 AUCH-Net\_{v/r} obtains better performance than Baseline. 
The above results show that the hypergraph networks, which learn high-order relationships between AU features and patch-level features, can effectively improve the performance.

\noindent \textbf{Effectiveness of the Relation Consistency Loss}
% 对比没有使用Relation Consistency Loss下的性能。
As shown in Table \ref{tab:ablation of AUCH-Net}, AUCH-Net\_a outperforms AUCH-Net\_{a/r} and AUCH-Net\_v obtains better performance than AUCH-Net\_{v/r} on both CFEE\_C  and RAF\_C datasets. 
These results show the effectiveness of the relation consistency loss. By
imposing the relation consistency loss on AU features or visual features extracted from hypergraphs, the model can learn consistent feature representations. Such a way is helpful to identify novel compound expression categories.

\begin{table}[!t]
\small
   \caption{Ablation study results on the AU regularization loss. `Intra' denotes the Intra-AU loss, `Inter' denotes the Inter-AU loss, and `Prior' denotes the Prior loss. The test accuracy (\%) of 5-way few-shot classification tasks with 95\% confidence intervals is reported. }
    \centering
    \renewcommand\arraystretch{1.1}{
    \resizebox{\linewidth}{!}{
	\setlength{\tabcolsep}{1.89mm}{
		\begin{tabular}{c c c | c c | c c}
			%\begin{tabular}{l|c|c|c|c}
			%\toprule [2 pt]
                  %\toprule
		\hline
        \multirow{2}{*}{Intra} & \multirow{2}{*}{Inter} & \multirow{2}{*}{Prior} & \multicolumn{2}{c}{CFEE\_C} & \multicolumn{2}{c}{RAF\_C} \\
	   & &  & 1-shot & 5-shot & 1-shot & 5-shot\\ 
        \hline
$\times$ & $\times$  & $\times$ & 55.98 \tiny{$\pm$ 0.71} & 68.21 \tiny{$\pm$ 0.56}& 46.77 \tiny{$\pm$ 0.63} & 61.92 \tiny{$\pm$ 0.63} \\
$\times$ & $\times$  & $\surd$ & 57.21 \tiny{$\pm$ 0.70} & 68.73 \tiny{$\pm$ 0.73}& 46.20 \tiny{$\pm$ 0.48} & 62.97 \tiny{$\pm$ 0.77} \\
$\surd$  & $\times$  & $\surd$ & 57.80 \tiny{$\pm$ 0.69} & 69.43 \tiny{$\pm$ 0.68}& 46.78 \tiny{$\pm$ 0.56} & 63.71 \tiny{$\pm$ 0.73} \\
$\times$ & $\surd $  & $\surd$ & 57.91 \tiny{$\pm$ 0.63} & 69.58 \tiny{$\pm$ 0.71}& 48.01 \tiny{$\pm$ 0.70} & 63.39 \tiny{$\pm$ 0.60} \\
$\surd$  & $\surd $  & $\surd$ & \textbf{58.36} \tiny{$\pm$ 0.37} & \textbf{70.11} \tiny{$\pm$ 0.74}& \textbf{48.65} \tiny{$\pm$ 0.68} & \textbf{64.55} \tiny{$\pm$ 0.49} \\
		\hline
                %\bottomrule
		\end{tabular} 
	}}
 }

 \label{tab:variants of ar}

\end{table}

\begin{comment}
\begin{table}[!t]
\small
    \caption{Ablation study results on the AU regularization loss. `Intra' denotes the Intra-AU loss, `Inter' denotes the Inter-AU loss, and `Prior' denotes the Prior loss. The test accuracy (\%) of 5-way few-shot classification tasks with 95\% confidence intervals is reported. }
    \centering
    \renewcommand\arraystretch{1.05}{
    \resizebox{\linewidth}{!}{
	\setlength{\tabcolsep}{1.89mm}{
		\begin{tabular}{c c c  c c  c c}
                  \toprule[1.5pt]
        \multirow{2}{*}{Intra} & \multirow{2}{*}{Inter} & \multirow{2}{*}{Prior} & \multicolumn{2}{c}{CFEE\_C} & \multicolumn{2}{c}{RAF\_C} \\
	   & &  & 1-shot & 5-shot & 1-shot & 5-shot\\ 
        \midrule[1.0pt]
            \ding{55}  & \ding{55}  & \ding{55}  & 55.98 & 68.21 & 46.77 & 61.92 \\
            \ding{55}  & \ding{55}  & \ding{52}  & 57.21 & 68.73 & 46.20 & 62.97 \\
            \ding{52}  & \ding{55}  & \ding{52}  & 57.80 & 69.43 & 46.78 & 63.71 \\
            \ding{55}  & \ding{52}  & \ding{52}  & 57.91 & 69.58 & 48.01 & 63.39 \\
            \ding{52}  & \ding{52}  & \ding{52}  & \textbf{58.36} & \textbf{70.11} & \textbf{48.65} & \textbf{63.90} \\
		
                \bottomrule[1.5pt]
		\end{tabular}
	}}
 }

 \label{tab:variants of ar}

\end{table}
\end{comment}

\begin{table}[!t]
\small
    \caption{Comparison of our inter-AU loss with the triplet loss and the contrastive loss. The Contrastive denotes the Contrastive loss and Triplet denotes the Triplet loss. The test accuracy (\%) of 5-way few-shot classification tasks with 95\% confidence intervals is reported.  }%}
    \centering
    \renewcommand\arraystretch{1.2}{
    \resizebox{\linewidth}{!}{
	\setlength{\tabcolsep}{1.89mm}{
		\begin{tabular}{l | c c | c c}
			%\begin{tabular}{l|c|c|c|c}
			%\toprule [2 pt]
                  %\toprule
		\hline
        \multirow{2}{*}{Method} & \multicolumn{2}{c}{CFEE\_C} & \multicolumn{2}{c}{RAF\_C} \\
	   & 1-shot & 5-shot & 1-shot & 5-shot\\ 
        \hline
    AUCH-Net\_{contrastive} & 56.92 \tiny{$\pm$ 0.46} & 68.31 \tiny{$\pm$ 0.71}& 47.93 \tiny{$\pm$ 0.71} & 62.10 \tiny{$\pm$ 0.59} \\
    AUCH-Net\_{triplet} & 57.11 \tiny{$\pm$ 0.58} & 68.69 \tiny{$\pm$ 0.69}& 47.32 \tiny{$\pm$ 0.74} & 62.66 \tiny{$\pm$ 0.68} \\
    AUCH-Net & \textbf{58.36} \tiny{$\pm$ 0.92} & \textbf{70.11} \tiny{$\pm$ 0.74}& \textbf{48.65} \tiny{$\pm$ 0.68} & \textbf{64.55} \tiny{$\pm$ 0.49} \\
		\hline
                %\bottomrule
		\end{tabular}
	}}
 }

 \label{tab:interau}

\end{table}

% \vspace{-11pt}

\noindent \textbf{Effectiveness of the AU Regularization Loss}
% 这边再插入一个对AU Regularization的消融实验
%The AU regularization loss consists of an intra-AU loss, an inter-AU loss, and a prior loss. 
To evaluate the importance of each loss in the AU regularization loss, we perform ablation studies on these losses. The results are shown in Table \ref{tab:variants of ar}. 

1) \noindent \textit{Effectiveness of the Prior Loss.}
Our method with only the prior loss achieves much better performance than our   method without the AU regularization loss. This shows the necessity of the prior loss, which enables the model to learn  appropriate AU features by leveraging a pretrained model.

2) \noindent \textit{Effectiveness of the Intra-AU Loss.} Our method with the   intra-AU and prior losses achieves better performance (0.68\% and 0.59\% improvements on the RAF\_C and CFEE\_C datasets, respectively, for the 5-way 1-shot task; 0.74\% and  0.70\% improvements on RAF\_C and CFEE\_C datasets, respectively, for the 5-way 5-shot task)  than that with only the prior loss. 
These results validate the effectiveness of the intra-AU loss, which emphasizes the relevant AUs for each expression category.

3) \noindent \textit{Effectiveness of the Inter-AU Loss.} Our method with  the inter-AU and the prior losses achieves better performance (1.21\% and 0.70\% improvements on the RAF\_C and CFEE\_C datasets, respectively, for the 5-way 1-shot task; 0.42\% and  0.85\% improvements on RAF\_C and CFEE\_C datasets, respectively, for the 5-way 5-shot task) than that with only the prior loss. The above results show the importance of the Inter-AU loss, which improves the compactness of AU features from the same category while enhancing the separability of AU features from the different categories. 

In addition,we also validate the superiority of the inter-AU loss against common losses (such as the triplet loss \cite{schroff2015facenet} and the contrastive loss \cite{hadsell2006dimensionality}). Specifically, we compare our AUCH-Net with AUCH-Net\_{contrastive} which replaces the inter-AU loss to the contrastive loss and AUCH-Net\_{triplet} which replaces the inter-AU loss to triplet loss.
We evaluate the performance 
on CFEE\_C dataset and RAF\_C dataset under the settings of 5-way 1-shot task and 5-way 5-shot task. 
We set the margin for the triplet loss to 0.5 and the margin for the contrastive loss to 0.5. The results are given in Table \ref{tab:interau}. 

We can see that our AUCH-Net obtains better performance than AUCH-Net\_{contrastive}  (1.44\% and 0.72\% improvements on the CFEE\_C and RAF\_C datasets, respectively, for the 5-way 1-shot task; 1.80\% and 1.80\% improvements on CFEE\_C and RAF\_C datasets, respectively, for the 5-way 5-shot task) and AUCH-Net\_{triplet}  (1.25\% and 1.33\% improvements on the CFEE\_C and RAF\_C datasets, respectively, for the 5-way 1-shot task; 1.42\% and 1.24\% improvements on CFEE\_C and RAF\_C datasets, respectively, for the 5-way 5-shot task). 
% 这个结果显示了我们的Inter-AU loss更好的让相同类别的AU相互靠近，不同类别的AU相互远离，更适用于我们的方法。
These results show that our proposed inter-AU loss can effectively pull AUs of the same category close while 
pushing AUs of different categories far away.

By combining the three losses, our method can achieve the best performance. The combination of three losses enables the model to learn discriminative AU features.

\begin{table*}[!t]
        \caption{Comparisons with state-of-the-art methods on the CFEE\_C, EmotioNet\_C, and RAF\_C datasets. The test accuracy (\%) of 5-way few-shot classification tasks with 95\% confidence intervals is reported.  The best and second-best results are marked in bold and underlined, respectively.}
    \renewcommand\arraystretch{0.96} % Reduced row height
    \scriptsize  % Smaller font size
    {
    \resizebox{\linewidth}{!}{
        \setlength{\tabcolsep}{1.5mm} % Decreased column separation
        \begin{tabular}{l| c | c c | c c | c c}
            \bottomrule[1.1pt]
            \multirow{2}{*}{Method} & \multirow{2}{*}{Venue} & \multicolumn{2}{c|}{CFEE\_C} & \multicolumn{2}{c|}{EmotioNet\_C} & \multicolumn{2}{c}{RAF\_C} \\
           & & 1-shot & 5-shot & 1-shot  & 5-shot & 1-shot & 5-shot \\
            \hline
            % \rowcolor{gray!12}
            \multicolumn{8}{c}{\textit{(a) Episodic training-based FSL methods}} \\
            \hline
            ProtoNet \cite{snell2017prototypical} & NeurIPS'2017 & 53.29 \tiny{$\pm$ 0.73} & 66.60  \tiny{$\pm$ 0.60} & 50.15 \tiny{$\pm$ 0.66} & 60.04 \tiny{$\pm$ 0.56} & 39.12 \tiny{$\pm$ 0.56} & 58.41 \tiny{$\pm$ 0.46} \\
            RelationNet \cite{sung2018learning} & CVPR'2018 & 50.58 \tiny{$\pm$ 0.68} & 63.17  \tiny{$\pm$ 0.60} & 48.33 \tiny{$\pm$ 0.68} & 56.27 \tiny{$\pm$ 0.58} & 36.18 \tiny{$\pm$ 0.54} & 53.45 \tiny{$\pm$ 0.46} \\
            GNN \cite{garcia2017few} & ICLR'2018 & 54.01 \tiny{$\pm$ 0.74} & 64.26  \tiny{$\pm$ 0.63} & 49.49 \tiny{$\pm$ 0.68} & 58.67 \tiny{$\pm$ 0.59} & 38.74 \tiny{$\pm$ 0.56} & 57.15 \tiny{$\pm$ 0.47} \\
            DSN \cite{simon2020adaptive} & CVPR'2020 & 49.61 \tiny{$\pm$ 0.73} & 60.03  \tiny{$\pm$ 0.62} & 48.25 \tiny{$\pm$ 0.68} & 54.89 \tiny{$\pm$ 0.58} & 40.09 \tiny{$\pm$ 0.55} & 52.49 \tiny{$\pm$ 0.47} \\
            InfoPatch \cite{liu2021learning} & AAAI'2021 & 54.19 \tiny{$\pm$ 0.67} & 67.29  \tiny{$\pm$ 0.56} & 48.14 \tiny{$\pm$ 0.61} & 59.84 \tiny{$\pm$ 0.55} & 41.02 \tiny{$\pm$ 0.52} & 57.98 \tiny{$\pm$ 0.45} \\
            CDKT \cite{NEURIPS2023_6cdb2cbb} & {NeurIPS'2023} & 55.70 \tiny{$\pm$ 0.66} & 68.56  \tiny{$\pm$ 0.69} & 51.68 \tiny{$\pm$ 0.56} & 60.13 \tiny{$\pm$ 0.73} & 43.76 \tiny{$\pm$ 0.68} & 59.91 \tiny{$\pm$ 0.59} \\
            \hline
            % \rowcolor{gray!12}
            \multicolumn{8}{c}{\textit{(b) Batch training-based FSL methods}} \\
            \hline
            Softmax \cite{chen2019closer} & ICML'2019 & 54.32 \tiny{$\pm$ 0.73} & 66.35  \tiny{$\pm$ 0.62} & 51.60 \tiny{$\pm$ 0.68} & 61.83 \tiny{$\pm$ 0.59} & 42.16 \tiny{$\pm$ 0.59} & 58.57 \tiny{$\pm$ 0.45} \\
            Cosmax \cite{chen2019closer} & ICML'2019 & 54.97 \tiny{$\pm$ 0.71} & 67.89  \tiny{$\pm$ 0.61} & 50.87 \tiny{$\pm$ 0.65} & 61.10 \tiny{$\pm$ 0.56} & 40.87 \tiny{$\pm$ 0.56} & 57.67 \tiny{$\pm$ 0.46} \\
            Arcmax \cite{afrasiyabi2020associative} & ECCV'2020 & 55.29 \tiny{$\pm$ 0.71} & 67.72  \tiny{$\pm$ 0.60} & 50.73 \tiny{$\pm$ 0.65} & 61.70 \tiny{$\pm$ 0.56} & 41.28 \tiny{$\pm$ 0.57} & 57.94 \tiny{$\pm$ 0.46} \\
            RFS \cite{tian2020rethinking} & ECCV'2020 & 55.21 \tiny{$\pm$ 0.70} & 66.45  \tiny{$\pm$ 0.71} & 52.07 \tiny{$\pm$ 0.59} & 61.89 \tiny{$\pm$ 0.44} & 43.31 \tiny{$\pm$ 0.68} & 61.11 \tiny{$\pm$ 0.72} \\
            LR+DC \cite{yang2021free} & {ICLR'2021} & 53.20 \tiny{$\pm$ 0.73} & 64.18  \tiny{$\pm$ 0.66} & 52.09 \tiny{$\pm$ 0.70} & 60.12 \tiny{$\pm$ 0.58} & 42.90 \tiny{$\pm$ 0.60} & 56.74 \tiny{$\pm$ 0.46} \\
            STARTUP \cite{phoo2020self} & {ICLR'2021} & 54.89 \tiny{$\pm$ 0.72} & 67.79  \tiny{$\pm$ 0.61} & 52.61 \tiny{$\pm$ 0.69} & 61.95 \tiny{$\pm$ 0.57} & 43.97 \tiny{$\pm$ 0.60} & 59.14 \tiny{$\pm$ 0.47} \\
            CDNet\_B \cite{zou2022learn} & ECCV'2022 & 54.55 \tiny{$\pm$ 0.71} & 68.09  \tiny{$\pm$ 0.62} & 52.76 \tiny{$\pm$ 0.67} & 61.76 \tiny{$\pm$ 0.57} & 42.02 \tiny{$\pm$ 0.58} & 61.75 \tiny{$\pm$ 0.44} \\
            HSM-Net \cite{chen2025hyperbolic} & TIP'2025 & \underline{57.96} \tiny{$\pm$ 0.86} & \underline{69.89}  \tiny{$\pm$ 0.70} & \underline{57.33} \tiny{$\pm$ 0.66} & \underline{64.95} \tiny{$\pm$ 0.58} & \underline{48.02} \tiny{$\pm$ 0.60} & \underline{64.23} \tiny{$\pm$ 0.45} \\
            \hline
            % \rowcolor{gray!12}
            \multicolumn{8}{c}{\textit{(c) Hybrid FSL methods}} \\
            \hline
            Meta-Baseline \cite{chen2021meta} & ICCV'2021 & 55.17 \tiny{$\pm$ 0.74} & 67.15  \tiny{$\pm$ 0.61} & 52.36 \tiny{$\pm$ 0.67} & 62.01 \tiny{$\pm$ 0.59} & 43.54 \tiny{$\pm$ 0.61} & 61.59 \tiny{$\pm$ 0.44} \\
            OAT \cite{chen2020diversity} & AAAI'2020 & 54.28 \tiny{$\pm$ 0.75} & 67.88  \tiny{$\pm$ 0.62} & 52.92 \tiny{$\pm$ 0.66} & 61.85 \tiny{$\pm$ 0.59} & 42.75 \tiny{$\pm$ 0.60} & 60.41 \tiny{$\pm$ 0.43} \\
            BML \cite{zhou2021binocular} & ICCV'2021 & 52.42 \tiny{$\pm$ 0.71} & 66.72  \tiny{$\pm$ 0.61} & 51.31 \tiny{$\pm$ 0.66} & 58.77 \tiny{$\pm$ 0.57} & 41.91 \tiny{$\pm$ 0.55} & 59.72 \tiny{$\pm$ 0.45} \\
            EGS-Net \cite{zou2022facial} & ICCV'2022 & 56.65 \tiny{$\pm$ 0.73} & 68.38  \tiny{$\pm$ 0.60} & 51.62 \tiny{$\pm$ 0.66} & 60.52 \tiny{$\pm$ 0.56} & 44.07 \tiny{$\pm$ 0.60} & 61.90 \tiny{$\pm$ 0.46} \\
            CDNet \cite{zou2022learn} & ECCV'2022 & 56.99 \tiny{$\pm$ 0.73} & 68.98  \tiny{$\pm$ 0.60} & 55.16 \tiny{$\pm$ 0.67} & 63.03 \tiny{$\pm$ 0.59} & 46.07 \tiny{$\pm$ 0.59} & 63.03 \tiny{$\pm$ 0.45} \\
            CPN \cite{lyu2023compositional} & AAAI'2023 & 56.73 \tiny{$\pm$ 0.74} & 68.45  \tiny{$\pm$ 0.70} & 52.60 \tiny{$\pm$ 0.45} & 59.83 \tiny{$\pm$ 0.62} & 44.65 \tiny{$\pm$ 0.55} & 62.31 \tiny{$\pm$ 0.39} \\
            LA-CMFER \cite{yang2024learning} & ACM MM'2024 & 56.59 \tiny{$\pm$ 0.67} & 67.91  \tiny{$\pm$ 0.77} & 56.01 \tiny{$\pm$ 0.70} & 63.42 \tiny{$\pm$ 0.49} & 45.72 \tiny{$\pm$ 0.48} & 62.83 \tiny{$\pm$ 0.56} \\

            \hline
            %\hline
            \rowcolor{gray!28}
            \textbf{AUCH-Net (Ours)} & - & \textbf{58.36} \tiny{$\pm$ 0.92} & \textbf{70.11}  \tiny{$\pm$ 0.74} & \textbf{57.60} \tiny{$\pm$ 0.64} & \textbf{65.29} \tiny{$\pm$ 0.66} & \textbf{48.65} \tiny{$\pm$ 0.68} & \textbf{64.55} \tiny{$\pm$ 0.49} \\ 
            \toprule[1.0pt]
        \end{tabular}
    }}

    \label{tab:compare}
\end{table*}

\begin{figure}[t!]
    \centering\includegraphics[width=0.48\textwidth]{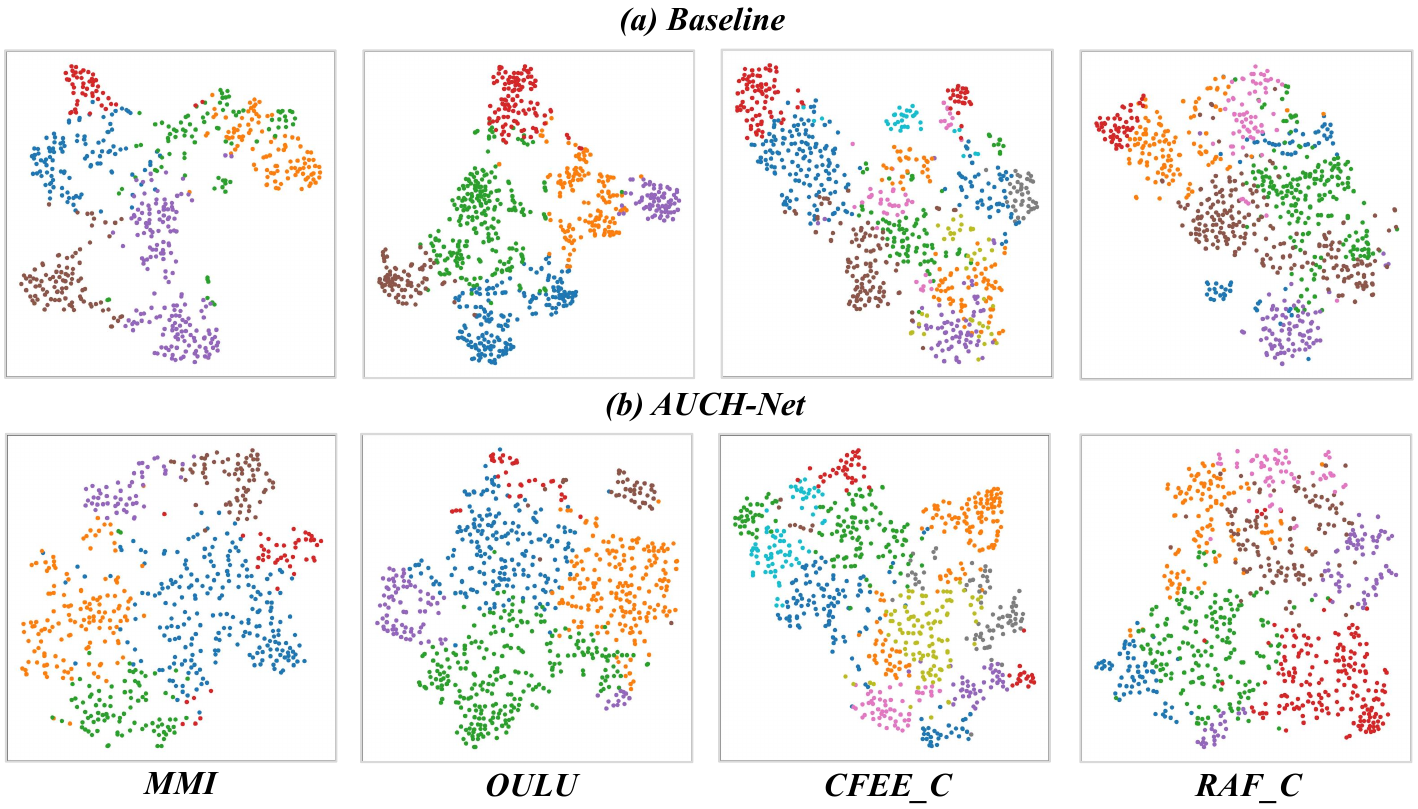}
    \caption{t-SNE visualization of the  feature obtained by (a) the Baseline method and (b) our AUCH-Net.
    The expression features of different datasets obtained by AUCH-Net have a consistent distribution, showing that AUCH-Net learns highly transferrable features.
    }
    \label{fig:visual3}
\end{figure}

\subsection{Visualization}
\label{visualization}
% 验证引入AU的作用，这边插入一张图，代表是否使用AU情况下对可视化划分数据集的作用。

%{Fig.~\ref{fig:heatmap} illustrates the attention heatmaps generated by the baseline method and our proposed AUCH-Net from original images and their flipped images across different datasets. It can be observed that 
%our method can learn more consistent attention heatmaps from the original images and their flipped images, thereby yielding a more robust model. This can be ascribed to the introduction of the novel relation consistency loss, which explicitly enforces the consistency between adjacency matrices from the original image and its flipped version. 

%In addition, 
We use t-SNE \cite{Maaten_Hinton_2008} to visualize the expression features extracted by the Baseline method and our AUCH-Net method on the source domain (including CK+, MMI, OULU) and the target domain (including CFEE\_C and RAF\_C). The results are shown in Fig.~\ref{fig:visual3}. We can observe that compared with the Baseline method, our AUCH-Net greatly reduces the domain discrepancy between the source and target domains. The expression features extracted from basic and compound
expressions are closely distributed, which validates that our method can learn a highly transferrable feature space.

\subsection{Computational Complexity}
\label{computation}
{
Our method takes lightweight module designs for multiple branches. The computational cost (FLOPs) and the number of parameters obtained by several competing methods are:  
{Baseline ({4.21G, 14.23M}), CDNet (4.78G, 13.23M), CPN (6.52G, 22.42M), and AUCH-Net (ours) (4.78G, 13.23M)}.
}  Hence, the computation complexity of our method is close to the Baseline method and state-of-the-art methods. 

\subsection{Comparison with State-of-the-Art Methods}
\label{SOTA}

To validate the effectiveness of our method, we compare our AUCH-Net with several state-of-the-art FSL methods (including episodic training-based, batch training-based, and hybrid FSL methods) on CFEE\_C,  EmotioNet\_C, and RAF\_C datasets under 1-shot and 5-shot settings. The results are given in Table \ref{tab:compare}.
%For a fair comparison, w
%We report the results obtained by the other competing methods using the source codes provided by respective authors under the same settings as ours.

As shown in Table \ref{tab:compare}, our  AUCH-Net achieves the best results on the CFEE\_C, RAF\_C, and EmotioNet\_C datasets for both the 5-way, 1-shot and 5-way, 5-shot classification tasks. Specifically, AUCH-Net achieves an accuracy of {58.36\%} on CFEE\_C, {57.60\%} on EmotioNet\_C, and {48.65\%} on RAF\_C for 5-way 1-shot classification tasks, and {70.11\%} on CFEE\_C, {65.29\%} on EmotioNet\_C, and {64.55\%} on RAF\_C for 5-way 5-shot classification tasks, respectively. 
% Compared with the batch training method HSM-Net, our method improve performance by {0.40\%} on CFEE\_C, on EmotioNet\_C, and {3.07\%} on RAF\_C for 5-way 1-shot classification tasks, and {0.83\%} on CFEE\_C, {2.05\%} on EmotioNet\_C, and {2.02\%} on RA\_C for 5-way 5-shot classification tasks. Moreover, compare with the episodic training-based method CDKT and hybrid training method CDNet, our method improve performance by {3.43\%} on CFEE\_C, {1.33\%} on EmotioNet\_C, and {5.37\%} on RAF\_C for 5-way 1-shot classification tasks, and {6.73\%} on CFEE\_C, {3.08\%} on EmotioNet\_C, and {3.56\%} on RAF\_C for 5-way 5-shot classification tasks. These results show the effectiveness of our proposed method.
Compared with the batch training method HSM-Net, our method improves performance by {0.40\%} on CFEE\_C, {0.27\%} on EmotioNet\_C, and {0.63\%} on RAF\_C for 5-way 1-shot classification tasks, and {0.22\%} on CFEE\_C, {0.34\%} on EmotioNet\_C, and {0.32\%} on RAF\_C for 5-way 5-shot classification tasks. Moreover, compared with the episodic training-based method CDKT and hybrid training method CDNet, our method achieves significant performance gains: against CDKT, we improve by {2.66\%} on CFEE\_C, {1.55\%} on EmotioNet\_C, and {5.92\%} on RAF\_C for 5-way 1-shot classification tasks, and {5.16\%} on CFEE\_C, {4.89\%} on EmotioNet\_C, and {4.64\%} on RAF\_C for 5-way 5-shot classification tasks; against CDNet, we outperform by {1.37\%} on CFEE\_C, {1.13\%} on EmotioNet\_C, and {2.44\%} on RAF\_C for 5-way 1-shot classification tasks, and {2.26\%} on CFEE\_C, {2.58\%} on EmotioNet\_C, and {1.52\%} on RAF\_C for 5-way 5-shot classification tasks. These results demonstrate the effectiveness of our proposed method.

The episodic training-based methods (such as ProtoNet  and RelationNet) easily suffer from the overfitting
problem caused by highly overlapped sampled tasks (since the base class set involves only a limited number of basic expressions in each few-shot task). Most batch training-based methods perform better than 
episodic training-based methods.
Existing hybrid FSL methods combine the two training paradigms to facilitate the training. Note that our method belongs to batch training-based methods, but it still achieves better performance than all hybrid FSL methods. Our method effectively models 
the connections between AUs and basic expressions in the source domain, facilitating learning consistent AU features in the target domain. As a result, our method can bridge the gap between expression images and expression categories, and learn {highly} transferable feature representations.

%As a result, our method can bridge the gap between expression images and expression categories, and learn {highly} transferable feature representations. 

\section{Conclusions}
\label{conclusion}

In this paper, we propose a novel AUCH-Net for CF-FER. AUCH-Net consists of three modules: a backbone, an AFL module, and a VFL module. Based on the basic features extracted from the backbone, the AFL module learns AU features by constructing consistency-aware hypergraphs, where a novel relation consistency loss and an AU regularization loss are developed. The VFL module learns visual features under the guidance of a relation consistency loss and a classification loss. By combining the outputs from the AFL and VFL modules, we can learn a highly  transferable expression feature space, which can be easily adapted to identify unseen compound expressions with very limited samples.  Qualitative and quantitative experiments consistently show the superiority of our AUCH-Net against several state-of-the-art methods for CF-FER.

Currently, our method shows promising performance by simply imposing the consistency between AUs corresponding to an input image and its flipped version on the hypergraphs. In future work, we will study other transforms to learn more consistent AU feature representations. 

\bibliographystyle{IEEEtran}
\bibliography{reference}

\end{document}